\pdfoutput=1
\PassOptionsToPackage{encapsulated}{CJK}
\documentclass[11pt]{article}
\usepackage{EMNLP2023}
\usepackage{times}
\usepackage{latexsym}
\usepackage[T1]{fontenc}
\usepackage[utf8]{inputenc}
\usepackage{microtype}
\usepackage{amsmath}
\usepackage{inconsolata}
\usepackage{microtype}
\usepackage{tikz}
\usepackage{float}
\usetikzlibrary{positioning, arrows.meta}  
\usepackage{inconsolata}
\usepackage{amsfonts} 
\usepackage{graphicx}
\usepackage{booktabs} 
\usepackage{algorithm}
\usepackage{algpseudocode}
\usepackage{multirow}
\usepackage{enumitem}
\usepackage{ruby}
\usepackage{CJKutf8}
\usepackage{booktabs}
\pagecolor{white} 
\usepackage{pifont}  
\usepackage{graphicx}
\usepackage{hyperref}   
\newcommand\blfootnote[1]{
  \begingroup
  \renewcommand\thefootnote{}\footnote{#1}
  \addtocounter{footnote}{-1}
  \endgroup
}
\newcommand{\cmark}{\textcolor{green!50!black}{\ding{51}}}
\newcommand{\xmark}{\textcolor{red!50!black}{\ding{55}}}
\setlist[enumerate]{nosep}
\setlist[itemize]{noitemsep} 

\def\framework{BIRD}

\title{BIRD: Bronze Inscription Restoration and Dating}

\author{
  Wenjie Hua \\
  Wuhan University \\
  \texttt{huawenjie@whu.edu.cn}
  \And
  Hoang H. Nguyen \\
  University of Illinois, Chicago \\
  \texttt{hnguy7@uic.edu}
  \And
  Gangyan Ge \\
  Xinjiang University \\
  \texttt{gegangyan@163.com}
}

\begin{document}
\maketitle

\begin{CJK}{UTF8}{bsmi} 
\begin{abstract}

Bronze inscriptions from early China are fragmentary and difficult to date. We introduce \textbf{BIRD} (\textbf{B}ronze \textbf{I}nscription \textbf{R}estoration and \textbf{D}ating), a fully encoded dataset grounded in standard scholarly transcriptions and chronological labels. We further propose an allograph-aware masked language modeling framework that integrates domain- and task-adaptive pretraining with a Glyph Net (GN), which links graphemes and allographs. Experiments show that GN improves restoration, while glyph-biased sampling yields gains in dating.

\end{abstract}

\blfootnote{%
  \begin{minipage}{0.95\linewidth} 
    \raggedright
    \raisebox{-0.2\height}{\includegraphics[height=1em]{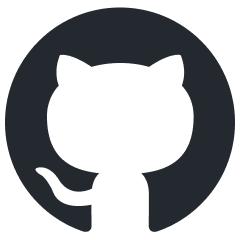}}~
    \href{https://github.com/wjhuah/BIRD}{\texttt{wjhuah/BIRD}} \\[0.3em]
  \end{minipage}
}

\section{Introduction}

Bronze inscriptions from the Chinese Bronze Age (21st–3rd c. BCE) are among the most important early Chinese textual sources \cite{lichuntao}. Found on ritual vessels, weapons, and musical instruments, these inscriptions record military achievements, feudal enfeoffments, oaths, and ancestral rites. Yet as excavated texts, they are often fragmentary and damaged, with uncertain chronological assignments.

\begin{figure}[t]
  \centering
  \includegraphics[width=0.95\linewidth]{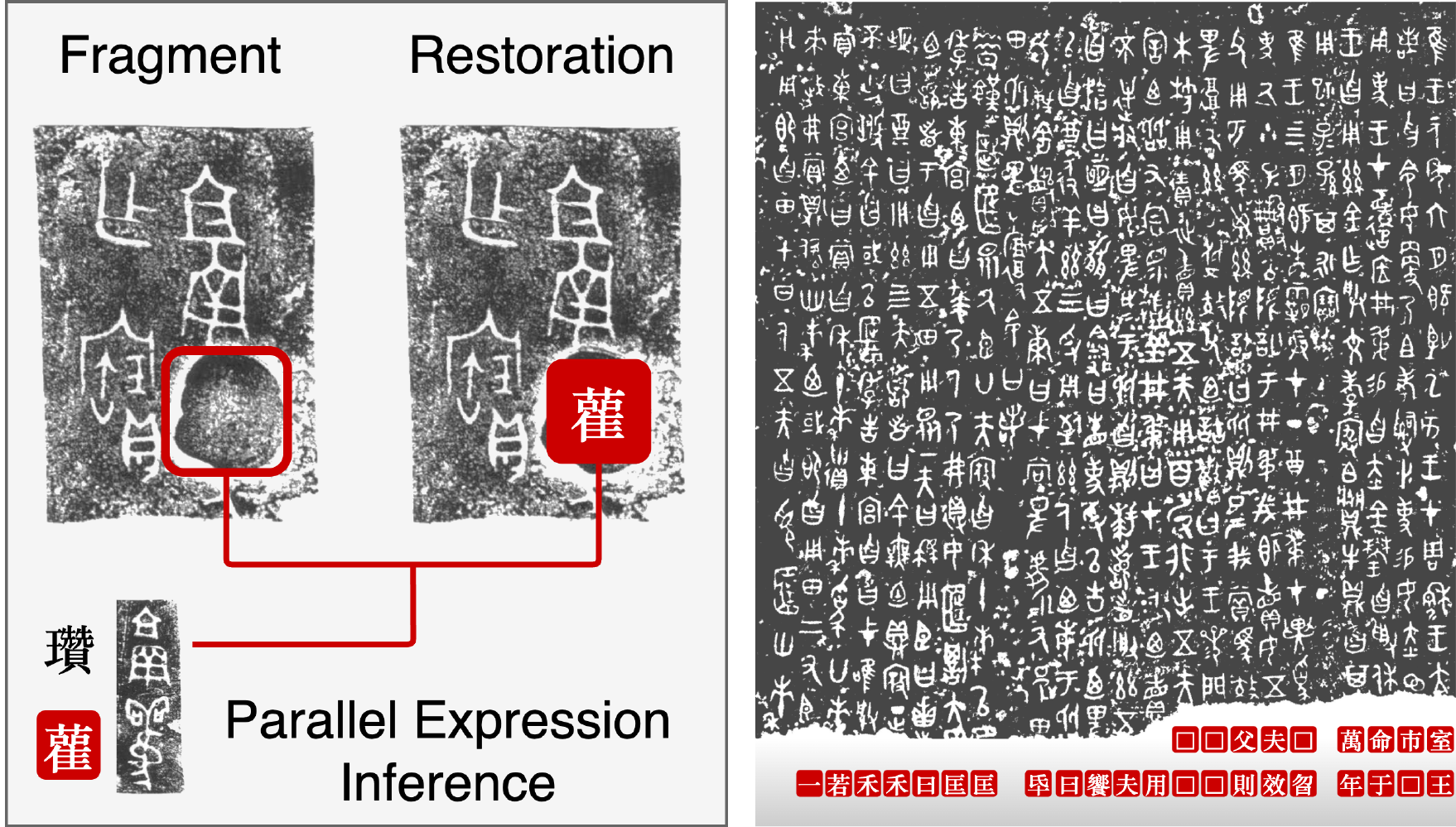}
  \caption{
  Left: A simplified paleographer’s workflow for restoring a damaged bronze inscription: identifying the damaged fragment, inferring from parallel expressions, and proposing a restoration 
  \cite{hubei1,wuzhenfeng,xiemingwen}. 
  Right: A damaged bronze inscription fragment (CCYZBI.02838A) \cite{cass} with the expert’s inferred reading \cite{huanghai}. The workflow mirrors a masked language modeling setup, where restorations are hypothesized from local context and attested parallel expressions.
  }
  \label{fig:fragment-example}
\end{figure}

Traditional restoration and dating rely on expert comparison of parallel expressions and contextual inference, along with other features, a process that is labor-intensive and difficult to scale. Neural models, particularly pre-trained language models (PLMs), have recently shown promise in supporting ancient text restoration. However, existing applications of artificial intelligence to bronze inscriptions focus almost exclusively on computer vision, such as single-character recognition or denoising of inscription images~\cite{zhinengjing,jinwenshibie1}. By contrast, natural language processing (NLP) approaches to inscriptional texts remain largely unexplored, despite their potential for tasks such as restoration and dating.

\begin{figure}[t]
  \centering
  \includegraphics[width=0.95\linewidth]{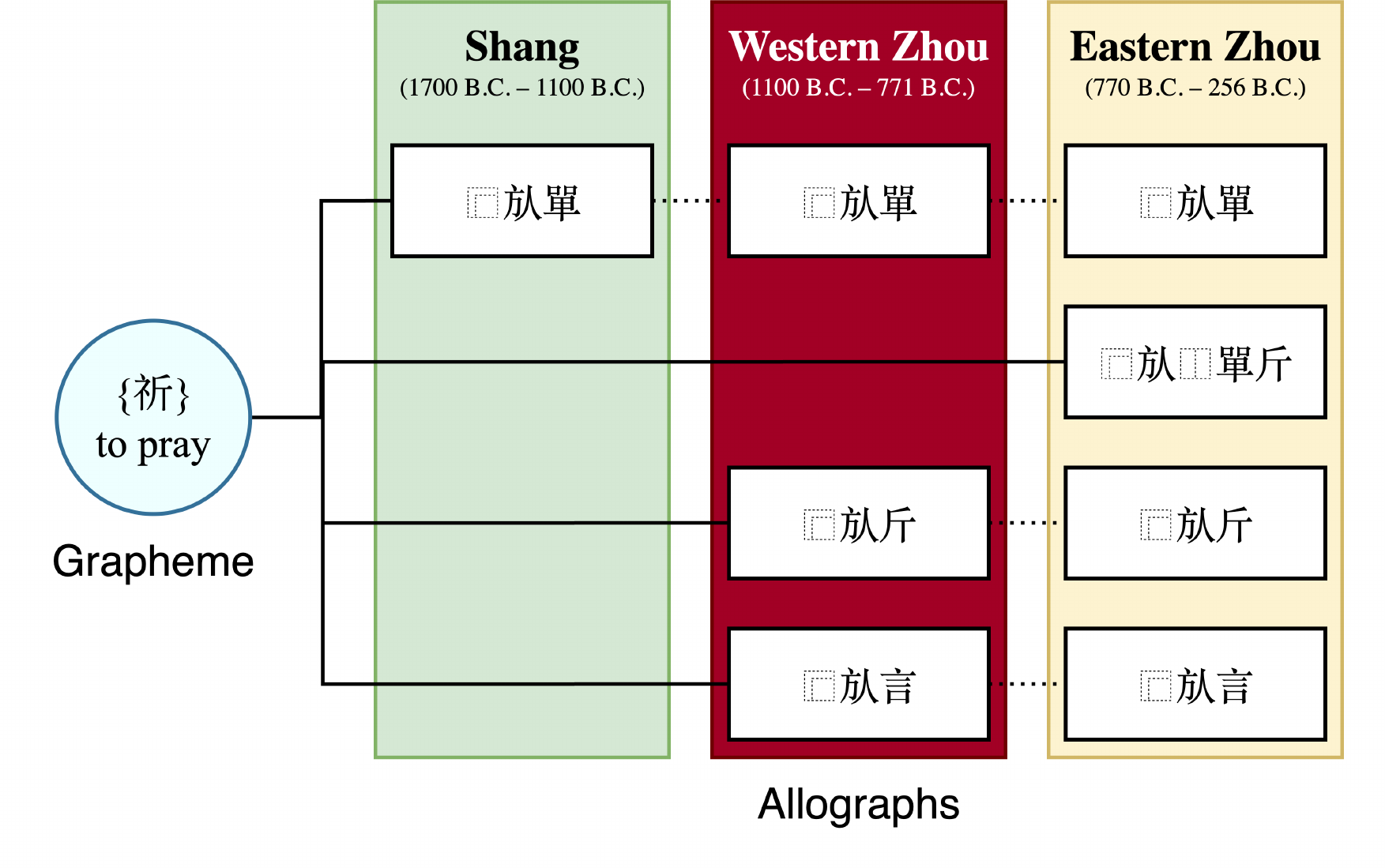}
  \caption{Concrete glyph family of \textit{Qi} (`to pray') from the Shang to the Eastern Zhou. To illustrate the correlation between glyphs and their components, Ideographic Description Sequences (IDS) are used.}
  \label{fig:gn-qi}
\end{figure}

Two factors make NLP modeling of bronze inscriptions challenging~\cite{lichuntao}: 

\begin{enumerate}[leftmargin=0.5cm]
    \item \textbf{Low-resource setting.} Although nearly 20k inscriptions have been published, most are extremely short, with over half containing three or fewer characters. Compared to modern corpora, the effective training data is therefore sparse.  

    \item \textbf{Allography.}\footnote{We use the term \textit{allograph} for distinct graphical forms that realize the same grapheme, following the graphematic perspective of ~\citet{types,me}. In Chinese palaeography, these correspond to so-called \textit{yitizi}. As ~\citet{qiuxigui1} notes, the broad definition of \textit{yitizi} subsumes two subtypes: narrow allographs (fully interchangeable forms) and partial allographs (forms that once overlapped in usage but later diverged, functionally close to \textit{tongyongzi}).} 
    In the Western Zhou corpus alone, 2,134 graphemes include 572 allograph sets (48.15\%) \cite{liuzhiji2}. Current encodings treat such forms as separate tokens, which prevents semantically equivalent allographs from being learned as a unified grapheme, thereby hindering generalization in data-hungry Transformers. Figure~\ref{fig:gn-qi} shows a representative family of allographs that share the same grapheme.

\end{enumerate}

Nonetheless, bronze inscriptions belong to the same synchronic register as transmitted and excavated Pre-Qin (21st–3rd c. BCE) texts \cite{lichuntao}, which can be leveraged as auxiliary data for domain-adaptive pretraining \cite{dont}. Moreover, allographic variation is not mere noise: in downstream tasks, e.g., chronological dating, glyph distinctions provide evidence \cite{yitiduandai1,gnxizhou1}. Effective modeling thus requires careful use of normalization to support low-resource learning, while still retaining the distinctive historical signal.

The contributions of this paper are as follows:
\vspace{-2pt}
\begin{itemize}
    \item \framework: the first fully encoded bronze inscription dataset (41k tokens) with encoding suitable for NLP tasks;
    \item The construction of Glyph Net (GN), a resource that pairs and clusters graphemes and their allographs into sets;  
    \item A framework for allograph-aware modeling that integrates GN for restoration and glyph-biased sampling for dating.   
\end{itemize}

\section{Related Work}
\label{relatedwork}

Research on bronze inscriptions has a long history. Paleographic resources, notably the widely accepted compilation of \citet{cass}, form the basis of \framework, complemented by the translations and philological studies of \citet{wuzhenfeng}, \citet{machengyuan}, and \citet{baichuanjing}. Chronological issues have also been examined, with frameworks proposed by \citet{tanglan}, \citet{chenmengjia}, and \citet{guomoruo}.

Digitization efforts have made significant contributions. The \textit{Digital Retrieval Platform for Shang and Zhou Bronze Inscriptions (Jihewang)} platform\footnote{\url{https://jwdcdbz.ancientbooks.cn}} integrates catalogs, glyph images, lexica, etc. Academia Sinica has released two semi-open databases: the \textit{Digital Archives of Bronze Images and Inscriptions (AS DABII)}\footnote{\url{https://bronze.asdc.sinica.edu.tw}}, covering vessel images, rubbings, typology, and metadata; and the \textit{Lexicon of Pre-Qin Oracle, Bronze Inscriptions and Bamboo Scripts (AS Lexicon)}\footnote{\url{https://inscription.asdc.sinica.edu.tw}}, spanning oracle bones, bronzes, and bamboo manuscripts for lexical research. However, these resources remain ill-suited for NLP tasks, as many characters, especially allographs, are represented only as images. Hence, addressing allography is crucial, with studies focusing on collecting examples across dynasties \citep{gnxizhou, gnshang, gnxizhou1, gndongzhou}.

Neural model restoration of fragmentary texts has been well-explored across languages. 
Most related to our work, \citet{mobofeng} applied BERT \citep{devlin2019bert} to masked character prediction on the Shanghai Museum bamboo manuscripts (1–9, 2,103 characters), simulating the speech case induction. 
\citet{beike} further combined RoBERTa \citep{roberta} with computer vision for restoring incomplete Chinese steles. 
In other low-resource epigraphic domains, similar approaches have achieved strong performance on Latin inscriptions \citep{lading}, Coptic manuscripts \citep{coptic}, Greek inscriptions \citep{assael2022}, and Akkadian cuneiform \citep{lazar2021}. 
Chronological dating tasks have also been pursued \citep{lading, chendanlu,iu}. 

Distinct from prior work, we provide the fully encoded and chronologically labeled bronze inscription corpus, accompanied by a grapheme-allograph resource, which enables neural models to tackle both restoration and dating.

\section{Dataset}
\vspace*{-2pt}
\subsection{Pre-Qin Corpus (DAPT)}
We perform domain-adaptive pretraining on Pre-Qin texts, covering 40 works across 11 categories with a total of 2.09M tokens, which were compiled from open corpora including the \textit{Chinese Text Project}\footnote{\url{https://ctext.org}} and Wikisource\footnote{\url{https://wikisource.org}}, and were further normalized (Appendix~\ref{preqin}).

\begin{table}[tbh]
\centering
\small
\begin{tabular}{lccccc}
\toprule
\textbf{Dataset} & \textbf{Ava.} & \textbf{Dedup.} & \textbf{Filt.} & \textbf{Enc.} & \textbf{Chron.} \\
\midrule
Jihewang         & \xmark & \xmark & \xmark & Partial & \cmark \\
AS DABII         & \xmark & \xmark & \xmark & Partial & \cmark \\
AS Lexicon       & \xmark & \xmark & \xmark & Partial & \cmark \\
\textbf{\framework}    & \cmark & \cmark & \cmark & Full    & \cmark \\
\bottomrule
\end{tabular}
\caption{Comparison of bronze inscription digitization efforts. Our dataset is the only publicly available, deduplicated, and filtered corpus, with complete encoding and chronological labels.}
\label{tab:digitization}
\vspace*{-3pt}
\footnotesize
\end{table}

\subsection{\framework{} (TAPT)}

Existing resources for bronze inscriptions, such as \textit{Jihewang}, \textit{AS DABII}, and \textit{AS Lexicon}, primarily serve as retrieval platforms rather than structured datasets. To fill this gap, we introduce \framework{}, the first NLP-ready dataset for inscription restoration and dating. 

\framework{} comprises 41k tokens and incorporates paleographic and digitization scholarship (Section~\ref{relatedwork}). Each inscription is annotated with the dynasty and finer-grained period, which supports supervised experiments on chronological classification. A comparison with previous digitization efforts is shown in Table~\ref{tab:digitization}.

In addition, we provide a Glyph Net resource of 1,078 grapheme–allograph pairs, compiled from Shang, Western Zhou, and Eastern Zhou studies~\cite{gnxizhou,gnshang,gndongzhou}. Following the principle of mutual substitutability~\cite{qiuxigui2}, it supports glyph family–level reasoning for MLM and downstream tasks.

\framework{} is prepared through four steps:  
\begin{enumerate}[leftmargin=0.5cm]
    \item \textbf{Encoding.} All inscriptions are converted into machine-readable text, with characters categorized into three types: identifiable, unreadable (\texttt{□}), and undeciphered (\texttt{[UNK]}; see Appendix~\ref{undiciphered}), as summarized in Table~\ref{tab:fuhao}.

    \item \textbf{Filtering.} Extremely short inscriptions ($\leq$1 character; 6,078 out of 17,547 in \textit{AS DABII}), mostly redundant single-character marks (e.g., ``Shi Ding'' consisting only of the character ``Shi,'' CCYZBI.01073--01088 \cite{cass}), are removed to avoid trivial patterns and ensure a more representative corpus.
    
    \item \textbf{Deduplication.} Many inscriptions recur across vessels (e.g., ten identical ``Bo Xian Fu Li,'' CCYZBI.00649--00658 \cite{cass}), as exact formulaic repetitions. Keeping all copies would inflate token counts, cause overfitting to duplicated patterns, and risk leakage, so we retain a single representative instance.

    \item \textbf{Correction.} Clerical transcriptions (\textit{liding}) and chronological assignments are revised according to recent philological studies, with efforts made to align the corpus with the most up-to-date interpretations. Further details are provided in Appendix~\ref{wholebib}.
\end{enumerate}

\begin{table}[t]
\centering
\begin{tabular}{lrr}
\toprule
\textbf{Type} & \textbf{Count} & \textbf{Proportion} \\
\midrule
Identifiable             & 39,565 & 99.24\% \\
Unreadable (\texttt{□})   & 236    & 0.59\%  \\
Undeciphered (\texttt{[UNK]}) & 56     & 0.14\%  \\
\bottomrule
\end{tabular}
\caption{Types of tokens and their proportions in \framework.}
\label{tab:fuhao}
\end{table}

\section{Model}
We use standard Transformer \citep{vaswani2017attention} masked-language-model (MLM) backbones, which have proven effective in text restoration tasks. Applying it to bronze inscriptions, however, presents two challenges: (i) the low-resource nature of the corpus, and (ii) the prevalence of allographs, where semantically equivalent forms appear as distinct tokens.  

\begin{figure}[t]
\centering
\includegraphics[width=0.95\linewidth]{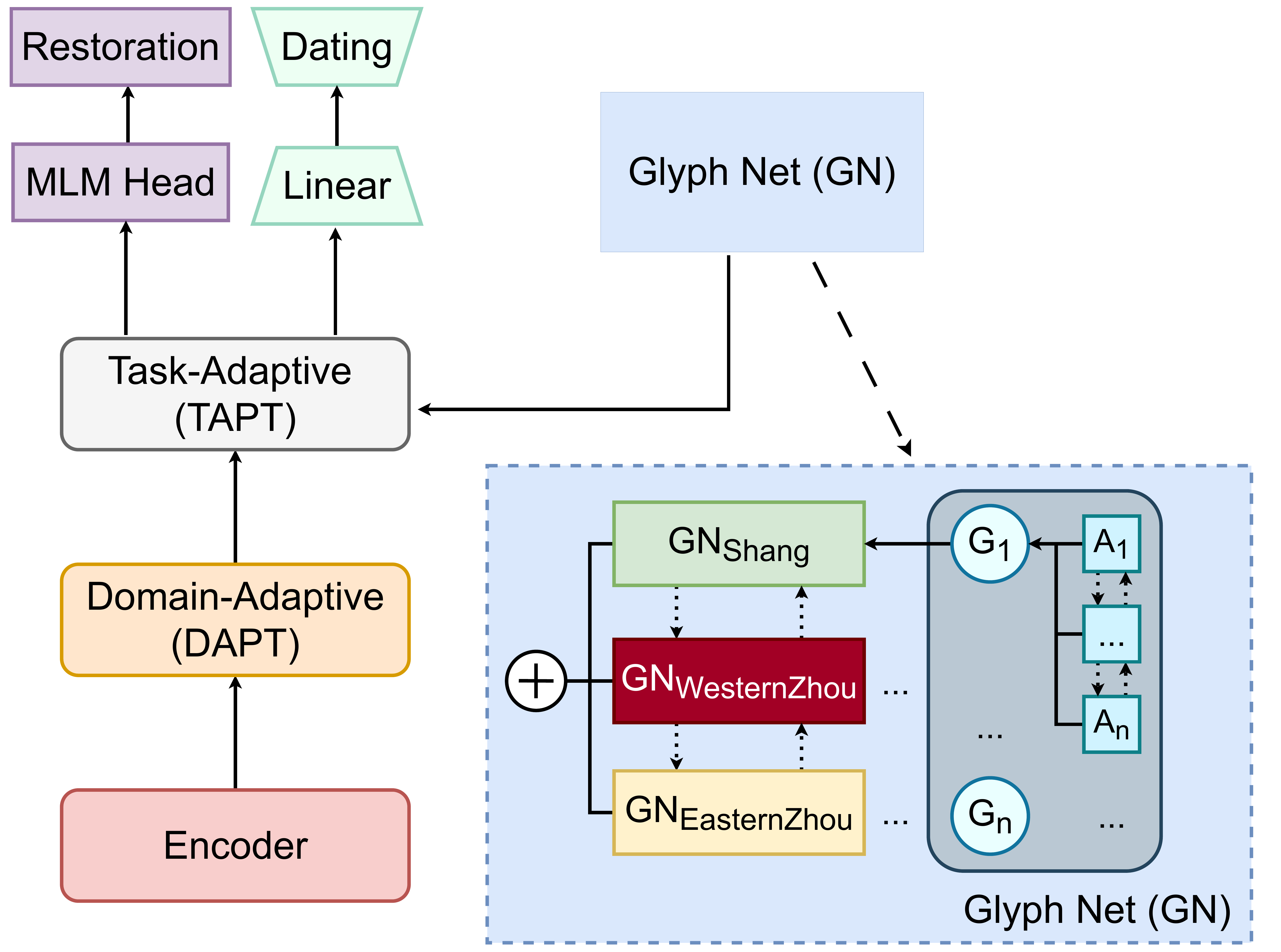}
\caption{Our pipeline enhances masked language modeling for bronze inscriptions by combining domain-adaptive pretraining (DAPT), task-adaptive pretraining (TAPT), and Glyph Net module (as illustrated in the lower-right component, each grapheme $G_{1..n}$ is linked to its allographs $A_{1..n}$) that integrates allograph glyph information into a BERT or RoBERTa backbone.
}
\label{fig:framework}
\end{figure}

To address these issues, we introduce three modifications to the MLM pipeline:
(1) domain-adaptive pretraining (DAPT) on a contemporaneous Pre-Qin corpus, with shallow layers frozen to stabilize training;  
(2) Glyph Net (GN), constructed from grapheme–allograph pairs. Transitive closure induces glyph families, and newly observed glyphs are aligned to family centroids;
(3) a glyph-biased sampling strategy that leverages GN families in two ways: aligning allographs for restoration, and emphasizing historically informative allographs for dating, inspired by weighted sampling techniques shown to enhance MLM training \cite{zhang2023bias}. Figure~\ref{fig:framework} illustrates the overall architecture.

\section{Experiments}\label{sec:experiments}
\vspace*{-2pt}
\subsection{Baselines}
We evaluate a BiLSTM sequence model as the restoration baseline \cite{lstm2, lstm1}, and an SVM classifier, which has shown strong performance in dynasty classification of historical Chinese texts \cite{iu}. 
For pretrained backbones, we consider \textsc{MultilingualBERT} (mBERT), \textsc{XLM-RoBERTa} (base and large) \cite{xlm}, and \textsc{SikuRoBERTa} \cite{sikubert}. 
Multilingual mBERT and XLM-R have demonstrated strong transfer performance in low-resource and cross-lingual settings \cite{lazar2021,multi}, 
while domain-specific models trained on the \textit{Siku Quanshu} corpus are widely adopted in ancient Chinese NLP \cite{2025eva,siku,mobofeng}.

\subsection{Implementation Details}
\label{masking}
Bronze inscriptions are extremely short, so standard BERT-style random masking often removes nearly all available context. 
To mitigate this, we adopt a stride-based masking scheme ($s$) that masks every $s$-th non-boundary character, so that sequences of length $\leq s$ lose at most one token. 
The stride parameter is tuned for each backbone using Bayesian hyperparameter search with Weights \& Biases \cite{wandb}.

\subsection{Tasks}
We model two complementary tasks that reflect real palaeographic challenges.  
For \textbf{restoration}, we apply the stride-based masking scheme (Section~\ref{masking}), and require the model to recover the gold character from incomplete inscriptions. Predictions are evaluated both at the exact character level and at the glyph-family level, where allographs under the same grapheme are treated as interchangeable.  

For \textbf{dating}, we fine-tune a linear head on the encoder representations to 
predict both dynasty-level and finer-grained period labels. The same backbone, settings, 
and adaptation schedules are shared across both tasks, which ensures comparability.

\section{Results and Discussion}
\vspace*{-2pt}
\subsection{Evaluation Criteria}

For \textbf{restoration}, we follow prior work \citep{assael2022, lazar2021} in single-position prediction. Performance is measured using:  
\emph{Exact@K}, which checks if the gold token appears within the top-$K$ predictions, and  
\emph{Family@K}, which counts a prediction correct if any member of the gold token’s allograph family appears within the top-$K$. All results are evaluated on glyph forms unseen during training, to approximate real restoration scenarios.

For \textbf{dating}, we evaluate at two granularities: dynasty-level (Shang, Western Zhou, Spring and Autumn, Warring States period), and period-level (Early, Middle, Late). We report accuracy and macro-F1, and additionally compute a hierarchical score that first verifies the dynasty label and then the period label within the predicted dynasty.

\subsection{Restoration Results}

Table~\ref{tab:resto-backbones} shows restoration results. 
\textsc{SikuRoBERTa} achieves the best performance on five of six metrics, including 49.47 Exact@1 and 73.07 Family@10, 
outperforming BiLSTM by +10.5\,p.p.\ (Exact@1) and +10.6\,p.p.\ (Family@10). 
BiLSTM leads on Family@1 (57.41). 
Multilingual PLMs lag \textsc{SikuRoBERTa} by 4–6\,p.p.\ on Exact@1, which confirms the advantage of in-domain pretraining. 

\begin{table}[t]
\centering
\small
\resizebox{\columnwidth}{!}{%
\begin{tabular}{lc|ccc|ccc}
\toprule
Model & Params & E@1 & E@5 & E@10 & F@1 & F@5 & F@10 \\
\midrule
BiLSTM            & 20M  & 39.02 & 42.98 & 53.10 & \textbf{57.41} & 57.63 & 62.50 \\
SikuRoBERTa & 109M & \textbf{49.47} & \textbf{65.20} & \textbf{70.15} & 54.32 & \textbf{68.05} & \textbf{73.07} \\
mBERT       & 110M & 43.55 & 58.57 & 63.71 & 46.93 & 61.28 & 65.92 \\
XLM-Base    & 278M & 43.51 & 58.35 & 62.94 & 44.28 & 59.49 & 64.03 \\
XLM-Large   & 550M & 45.64 & 60.92 & 64.91 & 47.16 & 61.17 & 65.36 \\
\bottomrule
\end{tabular}%
}
\caption{Restoration performance comparison of backbone models under the unified \textbf{GN} setting. 
\textbf{E@K} = Exact match at rank $K$; 
\textbf{F@K} = Family-level match at rank $K$.
All scores are percentages.}
\label{tab:resto-backbones}
\end{table}

\subsection{Dating Results}

Table~\ref{tab:dating-backbones} reports dynasty- and period-level dating. 
\textsc{SikuRoBERTa} delivers the best overall performance (dynasty accuracy 86.42; macro-F1 77.83) and the highest hierarchical dynasty accuracy (84.21). 
\textsc{mBERT} trails by 1–4 points, while larger multilingual encoders are less competitive. \textsc{XLM-Base} benefits slightly from period distinctions, and \textsc{XLM-Large} surpasses \textsc{SikuRoBERTa} on Hier-Per F1 but lags elsewhere. 
Overall, period dating proves more challenging than dynasty dating.

\begin{table}[t]
\centering
\small
\resizebox{\columnwidth}{!}{%
\begin{tabular}{lccccccc}
\toprule
\multirow{2}{*}{Model} & \multirow{2}{*}{Params} & \multicolumn{2}{c}{Dynasty} & \multicolumn{2}{c}{Hier-Dyn} & \multicolumn{2}{c}{Hier-Per} \\
\cmidrule(lr){3-4}\cmidrule(lr){5-6}\cmidrule(lr){7-8}
 &  & Acc & F1 & Acc & F1 & Acc & F1 \\
\midrule
SVM              & 0.08M   & 75.31 & 49.44 & 76.32 & 42.67 & 58.55 & 49.43 \\
SikuRoBERTa      & 109M  & \textbf{86.42} & \textbf{77.83} & \textbf{84.21} & \textbf{54.32} & \textbf{67.11} & 62.91 \\
mBERT            & 110M  & 84.57 & 74.77 & 82.24 & 53.12 & 63.82 & 58.63 \\
XLM-Base         & 278M  & 79.01 & 50.34 & 80.92 & 51.32 & 62.50 & 57.34 \\
XLM-Large        & 550M  & 84.01 & 74.60 & 81.58 & 53.12 & 65.13 & \textbf{62.96} \\
\bottomrule
\end{tabular}%
}
\caption{Dating performance comparison of backbones under the unified \textbf{glyph-biased} sampling. 
\textbf{Dynasty} = four-way classification; 
\textbf{Hier-Dyn} / \textbf{Hier-Per} = hierarchical evaluation at dynasty and period levels. 
All scores are percentages.}

\label{tab:dating-backbones}
\end{table}

\subsection{Analysis}

For \textbf{restoration}, GN is the most consistent contributor.
On the \textsc{SikuRoBERTa} backbone, GN achieves the highest Exact@K and Family@K (Table~\ref{tab:restoration-all}), with an average of 58.3 across metrics, surpassing Bias (57.8) and no adaptation (56.5), which demonstrates that collapsing allographs into glyph families effectively reduces sparsity and stabilizes restoration.  

For \textbf{dating}, the trend reverses.  
GN alone contributes little; instead, glyph-biased sampling toward glyph tokens yields the best dynasty- and period-level results (Table~\ref{tab:dyn-all}), averaging 68.9 across metrics—2.4 and 1.3 p.p.\ higher than GN (66.5) and GN+Bias (67.6), respectively. This empirically confirms that glyph distinctions serve as robust diachronic markers \cite{yitiduandai1,gnxizhou1}, while combining GN with bias adds no further gain. 

Takeaway: restoration benefits from modeling allographic equivalence, whereas dating exploits diachronic differentiation.

\subsection{Error Patterns}
Restoration is strongest in formulaic segments with fixed patterns \cite{machengyuan1}, while nouns denoting vessels, temporal adverbs, and modal particles achieve high accuracy due to their syntactic stability \cite{wuzhenyu}. Errors mainly default to frequent templates and confusions among semantically related nouns, verbs, or numerals, though even mispredictions often preserve syntactic category awareness. 

Misclassifications in dating concentrate in the Spring and Autumn and Warring States periods, where inscriptions are stylistically freer \cite{machengyuan1} and dynasty boundaries blur. Class imbalance also skews errors toward the Western Zhou. Nevertheless, severe cross-era errors remain rare, which indicates that the models effectively capture chronological signals.

\section{Conclusion}

We present \framework{} (\textbf{B}ronze \textbf{I}nscription \textbf{R}estoration and \textbf{D}ating), a curated dataset of transcribed and chronologically labeled bronze inscriptions.
On top of this dataset, we design a masked language modeling framework that integrates domain-adaptive pretraining (DAPT), task-adaptive pretraining (TAPT), and allograph-aware training.
Our experiments show that this framework, especially with the \textsc{SikuRoBERTa} backbone, achieves state-of-the-art results in both restoration and dating: Glyph Net (GN) stabilizes restoration, while glyph-biased sampling enhances chronological classification.
We hope BIRD provides a foundation for future NLP research on Chinese bronze inscriptions.

\section*{Limitations}

Despite promising gains in both restoration and dating, several limitations remain. First, \framework still suffers from sparsity and long-tail imbalance, which constrains generalization for rare forms. Related effort in this area can be found in \cite{longtail,nguyen2020dynamic} and similar studies.

Second, glyph-level modeling remains a challenge. Different characters may not consistently represent the same word \cite{qiuxigui1}, and our Glyph Net currently relies merely on inductive bias at the family level. Its generalization could be strengthened by incorporating stricter palaeographic constraints \cite{huangjuren} and expanding the knowledge base of loan characters \cite{tongjia}. Moreover, diachronic distributions of allographs are not well modeled, and the system is prone to semantically plausible but orthographically inappropriate predictions.

Third, our setup lacks phonological supervision. Bronze and other early Chinese inscriptions frequently employ loans ~\cite{baiyiping}, yet sound-based substitution is invisible to a token-only model. Incorporating phonetic series embeddings may capture such regularities, following phoneme-aware strategies that have proven effective in non-Latin scripts \cite{hoang1,hoang2}.

Fourth, fragmentary evidence poses a major obstacle. Some inscriptions are partially legible, with subcomponents visible even when the full graph is nearly damaged. A token-level MLM cannot leverage such partial signals. Structure-aware encodings, such as Ideographic Description Sequences (IDS), have been shown to be effective in related tasks \cite{ids1,ids2}, which could enable component-conditioned modeling for more robust restoration and dating.

Fifth, our framework omits archaeologically multimodal signals that are central to traditional dating. The shape of the vessel, the decorative motifs, and the casting techniques provide independent chronological evidence \cite{chenmengjia}, but are not yet explored. Integrating textual modeling with such modalities would bring the system closer to expert chronological practice.

Finally, our experiments are constrained by computational resources. 
We primarily relied on BERT- and RoBERTa-based backbones for sequence modeling due to their efficiency, 
whereas more recent generative architectures may better capture long-range dependencies and support free-form restoration. 
Exploring such models remains an open direction. 
Consequently, this work is positioned as a standardized dataset and a baseline framework to facilitate future research. 
Predictions should be viewed as auxiliary hypotheses, which offer preliminary guidance to paleographers,
with expert interpretation and archaeological context remaining indispensable.

\section*{Ethics Statement}
This work relies exclusively on ancient Chinese texts and bronze inscriptions, which contain no personal or sensitive information. The models are intended solely for academic research, and their predictions should not be regarded as authoritative readings of the inscriptions. We also acknowledge the environmental impact of model training, though our experiments involve relatively small-scale models with limited computational cost.

\section*{Acknowledgements}
We would like to thank Bin Li, Yuwen Zhang, Youzu He, Rongwei Yi, and the anonymous reviewers for their valuable feedback during the iterations of this paper. This project was supported in part by Wuhan University under the project “Research on Pre-Qin Inscriptions and Early Official Documents” (No. S202510486008).

\bibliography{anthology,custom}
\bibliographystyle{acl_natbib}

\newpage
\appendix

\section{Case Study: \textit{Hu Ding} Restoration}

To further illustrate our approach, we applied the model to the mid–Western Zhou \textit{Hu Ding} bronze vessel (CCYZBI.02838A/B), a well-known inscriptional source with multiple damaged positions. 
To avoid leakage, we excluded all \textit{Hu Ding} entries from \framework{} before training. 
The task is thus strictly out-of-sample.

\begin{figure}[ht]
\centering
\resizebox{\columnwidth}{!}{%
\includegraphics{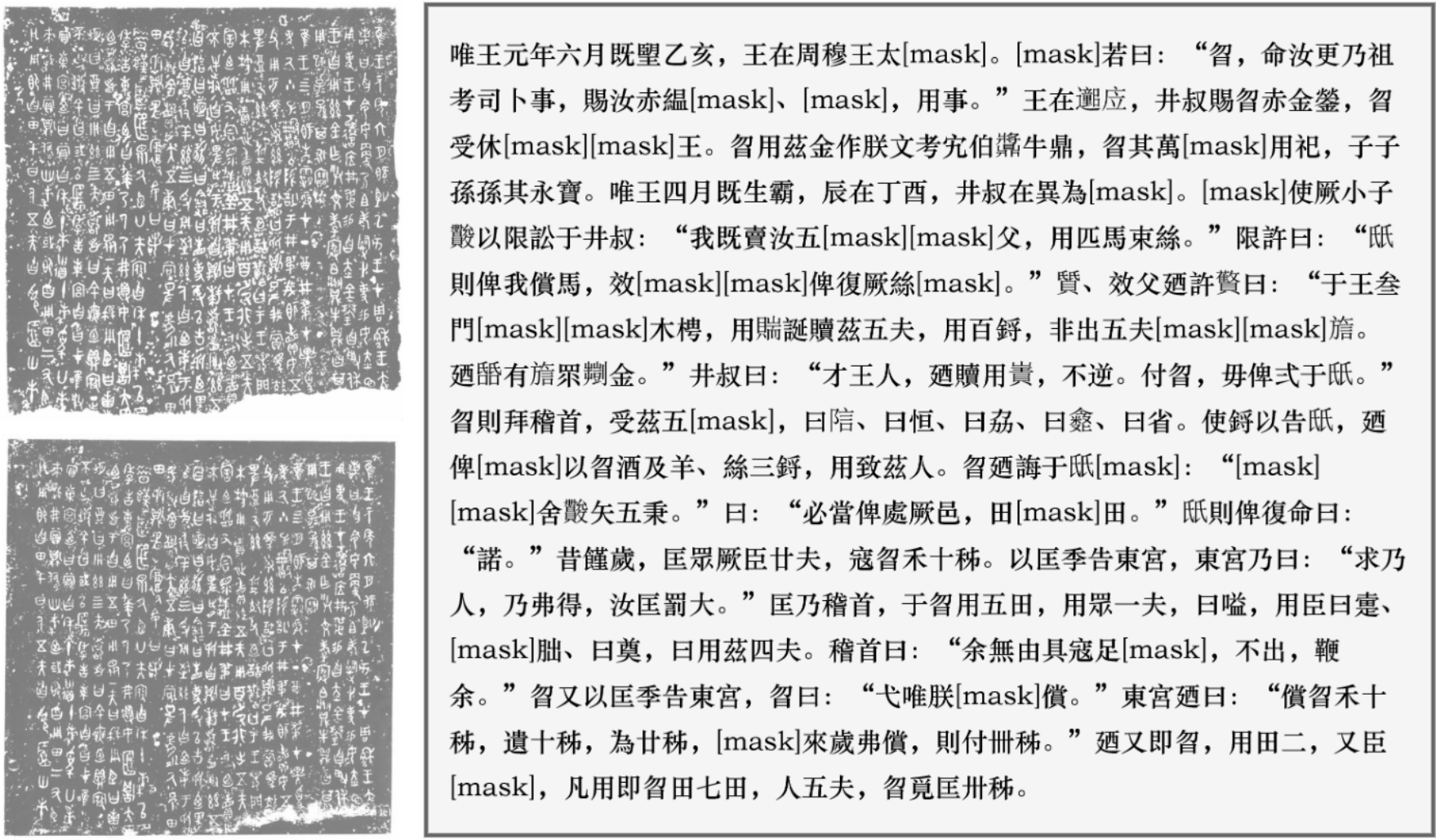}
}
\caption{
Left: Rubbing of the \textit{Hu Ding} inscriptions (CCYZBI.02838A, 02838B)~\cite{cass}, image courtesy of \textit{AS DABII}. 
Right: Transcription from~\cite{huanghai}, used as the input with damaged positions masked.}
\label{fig:hu-ding-inscription}
\end{figure}

We employed the SikuRoBERTa (GN) model with two decoding strategies: parallel mask filling and greedy iterative decoding~\cite{lazar2021}.
Table~\ref{tab:greedy-vs-gold} compares predicted tokens with expert gold restorations. 

\begin{table}[t]
\centering
\small
\begin{tabular}{c|c|c|c}
\toprule
\textbf{Mask Position} & \textbf{Gold} & \textbf{Pred@1} & \textbf{Top5} \\
\midrule
01 & 室 & 廟 & 廟 室 宮 寢 廷 \\
02 & 王 & 王 & 王 公 君 伯 尹 \\
03 & 芾 & 芾 & 芾 純 衡 衣 韍 \\
05 & 命 & 於 & 於 于 揚 無 多 \\
06 & 于 & 于 & 于 揚 穆 於 侑 \\
07 & 年 & 年 & 年 人 世 壽 歲 \\
\bottomrule
\end{tabular}
\caption{Top-1 and Top-5 predictions versus gold characters (excerpt of the first six damaged positions in the \textit{Hu Ding} inscription).}
\label{tab:greedy-vs-gold}
\end{table}

On 22 expert restorations~\cite{huanghai}, the model achieved Exact@1: 50.00\% (11/22), Exact@5: 59.09\% (13/22), and Exact@10: 68.18\% (15/22) under the parallel prediction setting. Greedy decoding yielded comparable coverage, though with a lower accuracy. In addition to reproducing expert restorations, the system generated plausible candidates for characters that remain undeciphered (Table~\ref{tab:undeciphered-results}), providing a potential reference for paleographic analysis.

\newcommand{\unicode}[1]{{\tiny #1}}

\begin{table}[t]
\centering
\small
\begin{tabular}{c|c}
\toprule
\textbf{Mask Position} & \textbf{Top10 Predictions} \\
\midrule
04 & 鑾 旂 舄 筆 \unicode{U+3AC3} 金 矢 黃 弓 璋 \\
08 & 介 伯 市 限 客 期 制 政 宰 人 \\
15 & 之 外 一 若 內 賜 邑 大 下 又 \\
16 & 賜 折 喬 杜 乘 造 擇 柞 之 于 \\
17 & 則 許 弗 不 人 亦 也 而 帛 乃 \\
18 & 則 曰 不 弗 許 告 厥 多 有 用 \\
28 & 其 厥 若 越 乃 我 以 汝 如 余 \\
\bottomrule
\end{tabular}
\caption{Model completions for undeciphered positions in \textit{Hu Ding} (Top-10 shown)}
\label{tab:undeciphered-results}
\end{table}

\section{DAPT Composition}
\label{preqin}

The DAPT (Pre-Qin) corpus consists of 40 transmitted and excavated texts, compiled and normalized from open sources. Following the classification of the \textit{Chinese Text Project}, Table~\ref{tab:dapt} presents a subset categorized. These texts provide broad coverage of syntactic and lexical patterns closely aligned with inscriptional Chinese.

\begin{table*}[t]
\centering
\small
\resizebox{\textwidth}{!}{%
\begin{tabular}{ll}
\toprule
\textbf{Category} & \textbf{Titles} \\
\midrule
Confucianism &
\textit{The Analects}, \textit{Mengzi}, \textit{Liji}, \textit{Xiao Jing}, \textit{Xunzi} \\
Mohism &
\textit{Mozi} \\
Daoism &
\textit{Dao De Jing}, \textit{Zhuangzi}, \textit{Liezi}, \textit{He Guan Zi}\\
Legalism &
\textit{Hanfeizi}, \textit{Shang Jun Shu}, \textit{Shenzi}, \textit{Jian Zhu Ke Shu}, \textit{Guanzi} \\
School of Names &
\textit{Gongsunlongzi} \\
School of the Military &
\textit{The Art of War}, \textit{Wu Zi}, \textit{Liu Tao}, \textit{Si Ma Fa}, \textit{Wei Liao Zi}  \\
Miscellaneous Schools &
\textit{Lü Shi Chun Qiu}, \textit{Gui Gu Zi} \\
Histories &
\textit{Chun Qiu Zuo Zhuan}, \textit{Lost Book of Zhou}, \textit{Guo Yu}, \textit{Yanzi Chun Qiu}, \textit{Zhan Guo Ce}, 
\textit{Zhushu Jinian}, \textit{Mutianzi Zhuan} \\
Ancient Classics &
\textit{Book of Poetry}, \textit{Shang Shu}, \textit{Book of Changes}, \textit{Rites of Zhou}, \textit{Chu Ci}, \textit{Yili}, \textit{Shan Hai Jing} \\
Medicine &
\textit{Huangdi Neijing} \\
Excavated &
\textit{Guodian}, \textit{Mawangdui} \\
\bottomrule
\end{tabular}}
\caption{Subset of Pre-Qin texts in the DAPT corpus. The Mawangdui Silk Texts, though excavated from a Western Han tomb (archaeological date), reflect Pre-Qin synchronic register (compositional date) and are thus included.}
\label{tab:dapt}
\end{table*}

\section{Training Objective}

Our model training combines three terms:

\paragraph{Masked language modeling (MLM).}
The base loss is the standard MLM objective:
\[
\mathcal{L}_{\text{MLM}} = - \sum_{i \in \mathcal{M}} \log P_\theta(y_i \mid X_i),
\]
where $\mathcal{M}$ is the set of masked positions, $y_i$ the gold glyph, and $X_i$ the masked context.

\paragraph{Glyph-biased masking.}
To emphasize historically informative glyphs, candidate positions are sampled with bias:
\[
p(i) = \frac{w_i}{\sum_{j \in \mathcal{C}} w_j}, \quad
w_i = 
\begin{cases}
\lambda & i \in \mathcal{G}, \\
1 & \text{otherwise},
\end{cases}
\]
where $\mathcal{C}$ is the set of candidate tokens, $\mathcal{G}$ the set of glyph tokens in GN clusters, and $\lambda>1$ controls the bias strength.

\paragraph{Glyph-net (GN) regularization.}
To encourage consistency across allographs, for each cluster $G$ with token set $\mathcal{T}(G)$ we define:
\[
\mathcal{L}_{\text{GN}} = - \frac{1}{|\mathcal{M}|} \sum_{i \in \mathcal{M}}
 \frac{1}{|\mathcal{T}(G_i)|} \sum_{t \in \mathcal{T}(G_i)} \log P_\theta(t \mid X_i).
\]

\paragraph{Final objective.}
The training loss interpolates between MLM and GN terms:
\[
\mathcal{L} = (1-\alpha)\,\mathcal{L}_{\text{MLM}} + \alpha\,\mathcal{L}_{\text{GN}},
\]
with $\alpha$ gradually scheduled during training.

\newpage

\section{Training Setup} 

We evaluate four adaptation schedules: (i) no adaptation, (ii) domain-adaptive pretraining (DAPT), (iii) task-adaptive pretraining (TAPT), and (iv) a two-stage DAPT→TAPT pipeline. Each schedule is optionally combined with Glyph Net alignment or glyph-biased sampling. All baselines are extended with UNK placeholders for unseen characters. In DAPT, the bottom six layers are frozen and trained for ten epochs on a large Pre-Qin corpus; TAPT then unfreezes all layers and adapts to inscriptional data. The two stages are interpolated with a weighting parameter~$\lambda$ balancing DAPT and TAPT losses.

\section{Hyper-parameters}

We found the best hyperparameters for each model during the search via Weights \& Biases \cite{wandb}, as detailed in \autoref{tab:hyperparam-xlm}.

\begin{table}[htb]
    \centering
    \small
    \resizebox{\columnwidth}{!}{%
    \begin{tabular}{|l|l|l|l|l|}
        \hline
        \textbf{Hyper-parameter} & \textbf{mBERT} & \textbf{XLM-Base} & \textbf{XLM-Large} & \textbf{SikuRoBERTa} \\
        \hline
        Learning Rate & 0.00005 & 0.00005 & 0.00005 & 0.00012 \\
        \hline
        Epochs & 60 & 40 & 40 & 40 \\
        \hline
        Batch Size & 32 & 32 & 32 & 32 \\
        \hline
        Attention Dropout & 0.1 & 0.1 & 0.1 & 0.1 \\
        \hline
        Hidden Dropout & 0.1 & 0.1 & 0.1 & 0.1 \\
        \hline
        Stride & 12 & 10 & 12 & 10 \\
        \hline
        mlm\_prob & 0.2 & 0.2 & 0.2 & 0.2 \\
        \hline
        Weight Decay & 0.01 & 0.01 & 0.01 & 0.01 \\
        \hline
    \end{tabular}%
    }
    \caption{Best hyperparameters found during WandB hyperparameter search for mBERT, XLM-Base, XLM-Large, and SikuRoBERTa.}
    \label{tab:hyperparam-xlm}
\end{table}

\section{Undeciphered Characters}
\label{undiciphered}
\begin{figure}[t]
\centering
\setlength{\tabcolsep}{3pt}
\renewcommand{\arraystretch}{1.2}
\begin{tabular}{cccccc}
	\includegraphics[width=0.12\linewidth]{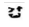} &
	\includegraphics[width=0.12\linewidth]{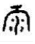} &
	\includegraphics[width=0.12\linewidth]{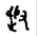} &
	\includegraphics[width=0.12\linewidth]{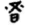} &
	\includegraphics[width=0.12\linewidth]{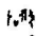} &
	\includegraphics[width=0.12\linewidth]{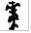} \\
	
	\includegraphics[width=0.12\linewidth]{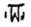} &
	\includegraphics[width=0.12\linewidth]{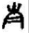} &
	\includegraphics[width=0.12\linewidth]{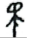} &
	\includegraphics[width=0.12\linewidth]{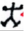} &
	\includegraphics[width=0.12\linewidth]{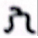} &
	\includegraphics[width=0.12\linewidth]{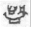} \\
	
	\includegraphics[width=0.12\linewidth]{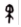} &
	\includegraphics[width=0.12\linewidth]{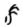} &
	\includegraphics[width=0.12\linewidth]{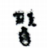} &
	\includegraphics[width=0.12\linewidth]{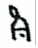} &
	& \\
\end{tabular}

\caption{Examples of undeciphered glyphs represented by UNK placeholders in \framework.}
\label{fig:unk-glyphs}
\end{figure}

Figure~\ref{fig:unk-glyphs} shows glyphs from bronze inscriptions that remain undeciphered by paleographers. The complete collection of undeciphered forms can be found in our GitHub repository. We encode these glyphs with symbolic placeholders, assigning a distinct identifier to each form. Visually identical forms are mapped to the same identifier.

\section{Paleographical References}\label{wholebib}
We draw on recent paleographical and historical studies of bronze inscriptions to update character forms and chronological assignments in our corpus. 
For example, an inscription previously dated to the Early Spring and Autumn period (CCYZBI.02737 \cite{cass}) has been reassigned to the Middle Spring and Autumn period in \cite{wuzhenfeng}; similarly, another item formerly placed in the Middle Western Zhou (CCYZBI.02737 \cite{cass}) has been revised to the Middle Spring and Autumn period in \cite{fanguo}. 
For further details, please refer to our GitHub repository, which will continue to be updated in the future.

\section{Ablation}

We conduct ablation studies across four backbones (\textsc{SikuRoBERTa}, \textsc{mBERT}, \textsc{XLM-Base}, \textsc{XLM-Large}) to disentangle the effects of domain- and task-adaptive pretraining, allograph-aware supervision, and glyph-biased sampling. 
Restoration accuracy is summarized in Table~\ref{tab:restoration-all}, dating performance is reported in Table~\ref{tab:dyn-all}, and representation cohesion and separation are analyzed in Table~\ref{tab:representation-all}.

\clearpage

\clearpage
\begin{table*}[t]
\centering
\small
\begin{tabular}{l lcccccc}
\toprule
Model & Scenario & E@1 $\uparrow$ & E@5 $\uparrow$ & E@10 $\uparrow$ & F@1 $\uparrow$ & F@5 $\uparrow$ & F@10 $\uparrow$ \\
\midrule
\multirow{7}{*}{\textsc{SikuRoBERTa}}
& Baseline        & 0.236 & 0.377 & 0.440 & 0.244 & 0.395 & 0.458 \\
& DAPT\_only      & 0.260 & 0.423 & 0.494 & 0.253 & 0.432 & 0.512 \\
& TAPT\_Bias      & 0.483 & 0.626 & 0.676 & 0.544 & 0.678 & 0.731 \\
& TAPT\_GN        & \textbf{0.495} & \textbf{0.652} & \textbf{0.702} & 0.543 & 0.681 & \textbf{0.731} \\
& TAPT\_GN\_Bias  & 0.492 & 0.638 & 0.686 & \textbf{0.554} & \textbf{0.688} & 0.729 \\
& TAPT\_from\_DAPT& 0.485 & 0.636 & 0.685 & 0.535 & 0.681 & 0.729 \\
& TAPT\_only      & 0.488 & 0.639 & 0.684 & 0.539 & 0.681 & 0.723 \\
\midrule
\multirow{7}{*}{\textsc{mBERT}}
& Baseline        & 0.112 & 0.224 & 0.282 & 0.093 & 0.205 & 0.267 \\
& DAPT\_only      & 0.148 & 0.283 & 0.353 & 0.139 & 0.278 & 0.355 \\
& TAPT\_Bias      & 0.427 & 0.572 & 0.622 & 0.464 & 0.617 & 0.665 \\
& TAPT\_GN        & \textbf{0.436} & \textbf{0.586} & \textbf{0.637} & \textbf{0.469} & 0.613 & 0.659 \\
& TAPT\_GN\_Bias  & 0.424 & 0.583 & 0.635 & 0.466 & \textbf{0.618} & \textbf{0.665} \\
& TAPT\_from\_DAPT& 0.431 & 0.574 & 0.623 & 0.464 & 0.607 & 0.657 \\
& TAPT\_only      & 0.427 & 0.570 & 0.613 & 0.465 & 0.606 & 0.648 \\
\midrule
\multirow{7}{*}{\textsc{XLM-Base}}
& Baseline        & 0.122 & 0.195 & 0.234 & 0.112 & 0.187 & 0.228 \\
& DAPT\_only      & 0.161 & 0.279 & 0.337 & 0.151 & 0.270 & 0.332 \\
& TAPT\_Bias      & 0.432 & 0.572 & 0.622 & \textbf{0.454} & 0.598 & 0.644 \\
& TAPT\_GN        & \textbf{0.435} & \textbf{0.584} & \textbf{0.629} & 0.443 & 0.595 & 0.640 \\
& TAPT\_GN\_Bias  & 0.429 & 0.583 & 0.626 & \textbf{0.454} & \textbf{0.608} & \textbf{0.651} \\
& TAPT\_from\_DAPT& 0.434 & 0.583 & 0.629 & 0.447 & 0.595 & 0.639 \\
& TAPT\_only      & 0.424 & 0.557 & 0.602 & 0.434 & 0.568 & 0.614 \\
\midrule
\multirow{7}{*}{\textsc{XLM-Large}}
& Baseline        & 0.140 & 0.225 & 0.265 & 0.132 & 0.208 & 0.257 \\
& DAPT\_only      & 0.178 & 0.321 & 0.382 & 0.166 & 0.312 & 0.384 \\
& TAPT\_Bias      & 0.453 & 0.595 & 0.640 & \textbf{0.479} & \textbf{0.615} & 0.656 \\
& TAPT\_GN        & \textbf{0.456} & \textbf{0.609} & \textbf{0.649} & 0.472 & 0.612 & 0.654 \\
& TAPT\_GN\_Bias  & 0.454 & 0.598 & 0.645 & 0.476 & 0.609 & \textbf{0.657} \\
& TAPT\_from\_DAPT& 0.456 & 0.600 & 0.648 & 0.471 & 0.604 & 0.649 \\
& TAPT\_only      & 0.435 & 0.577 & 0.621 & 0.442 & 0.584 & 0.622 \\
\bottomrule
\end{tabular}
\caption{Restoration results under different adaptation schedules across four pretrained models. 
E@k denotes Exact@k and F@k denotes Family@k. 
All values are reported as proportions between 0 and 1. 
Best results per column are bolded.}

\label{tab:restoration-all}
\end{table*}

\clearpage

\begin{table*}[p]
\centering
\small
\resizebox{\textwidth}{!}{%
\begin{tabular}{l lcccccc}
\toprule
Model & Scenario & Acc\_Dyn $\uparrow$ & F1\_Dyn $\uparrow$ & Acc\_Hier\_Dyn $\uparrow$ & F1\_Hier\_Dyn $\uparrow$ & Acc\_Hier\_Per $\uparrow$ & F1\_Hier\_Per $\uparrow$ \\
\midrule
\multirow{6}{*}{\textsc{SikuRoBERTa}} 
& DAPT\_only       & 0.833 & 0.728 & 0.836 & \textbf{0.552} & \textbf{0.684} & 0.630 \\
& TAPT\_only       & 0.846 & 0.727 & \textbf{0.849} & 0.570 & 0.651 & 0.605 \\
& TAPT\_from\_DAPT & 0.840 & 0.698 & 0.836 & 0.544 & 0.671 & 0.627 \\
& TAPT\_GN         & 0.840 & 0.698 & 0.842 & 0.542 & \textbf{0.684} & \textbf{0.638} \\
& TAPT\_GN\_Bias   & 0.852 & 0.767 & 0.849 & 0.539 & 0.678 & 0.635 \\
& TAPT\_Bias       & \textbf{0.864} & \textbf{0.778} & 0.842 & 0.543 & 0.671 & 0.629 \\
\midrule
\multirow{6}{*}{\textsc{mBERT}} 
& Baseline         & 0.809 & 0.672 & 0.822 & 0.515 & 0.638 & 0.583 \\
& TAPT\_only       & \textbf{0.846} & \textbf{0.762} & \textbf{0.836} & 0.540 & \textbf{0.664} & \textbf{0.616} \\
& TAPT\_from\_DAPT & \textbf{0.846} & 0.752 & 0.822 & 0.534 & 0.658 & \textbf{0.616} \\
& TAPT\_GN         & \textbf{0.846} & 0.745 & 0.822 & 0.534 & 0.618 & 0.575 \\
& TAPT\_GN\_Bias   & 0.815 & 0.642 & 0.829 & \textbf{0.547} & 0.651 & 0.601 \\
& TAPT\_Bias       & \textbf{0.846} & 0.748 & 0.822 & 0.531 & 0.638 & 0.586 \\
\midrule
\multirow{7}{*}{\textsc{XLM-Base}} 
& Baseline         & 0.673 & 0.277 & 0.763 & 0.379 & 0.592 & 0.520 \\
& DAPT\_only       & 0.778 & 0.483 & 0.796 & 0.493 & 0.625 & 0.567 \\
& TAPT\_only       & 0.778 & 0.476 & 0.789 & 0.498 & 0.612 & 0.566 \\
& TAPT\_from\_DAPT & 0.784 & 0.486 & \textbf{0.809} & 0.484 & 0.618 & \textbf{0.576} \\
& TAPT\_GN         & 0.765 & 0.429 & 0.803 & 0.433 & \textbf{0.632} & 0.569 \\
& TAPT\_GN\_Bias   & 0.784 & 0.481 & \textbf{0.809} & 0.502 & 0.605 & 0.556 \\
& TAPT\_Bias       & \textbf{0.790} & \textbf{0.503} & \textbf{0.809} & \textbf{0.513} & 0.625 & 0.573 \\
\midrule
\multirow{7}{*}{\textsc{XLM-Large}} 
& Baseline         & 0.747 & 0.444 & 0.803 & 0.436 & 0.618 & 0.542 \\
& DAPT\_only       & 0.772 & 0.566 & 0.822 & 0.540 & 0.612 & 0.580 \\
& TAPT\_only       & 0.815 & 0.667 & \textbf{0.849} & 0.572 & \textbf{0.678} & \textbf{0.657} \\
& TAPT\_from\_DAPT & 0.809 & 0.655 & \textbf{0.849} & \textbf{0.581} & 0.658 & 0.630 \\
& TAPT\_GN         & 0.821 & 0.705 & 0.809 & 0.512 & 0.572 & 0.534 \\
& TAPT\_GN\_Bias   & \textbf{0.840} & 0.701 & 0.836 & 0.526 & 0.612 & 0.561 \\
& TAPT\_Bias       & \textbf{0.840} & \textbf{0.746} & 0.816 & 0.531 & 0.651 & 0.630 \\
\bottomrule
\end{tabular}%
}
\caption{Classification results for dynasty- and period-level dating under different adaptation schedules. 
Acc = accuracy, F1 = macro-F1. 
\textbf{Dyn} = dynasty-level classification (single-task); 
\textbf{Hier\_Dyn} = dynasty-level accuracy/F1 in the hierarchical model; 
\textbf{Hier\_Per} = period-level accuracy/F1 in the hierarchical model, where period prediction is conditioned on the predicted dynasty. 
Best results per column are bolded.}
\label{tab:dyn-all}
\end{table*}

\clearpage

\begin{table*}[t]
\centering
\small
\begin{tabular}{l lcc}
\toprule
Model & Scenario & IntraCos Avg ($\uparrow$) & Nearest-InterCos Avg ($\downarrow$) \\
\midrule
\multirow{7}{*}{\textsc{SikuRoBERTa}}
& Baseline        & 0.494 & 0.252 \\
& DAPT\_only      & 0.488 & 0.266 \\
& TAPT\_Bias      & 0.503 & 0.292 \\
& TAPT\_GN        & \textbf{0.515} & 0.291 \\
& TAPT\_GN\_Bias  & 0.514 & 0.290 \\
& TAPT\_from\_DAPT& 0.504 & 0.295 \\
& TAPT\_only      & 0.486 & \textbf{0.255} \\
\midrule
\multirow{7}{*}{\textsc{mBERT}}
& Baseline        & 0.496 & 0.218 \\
& DAPT\_only      & 0.470 & \textbf{0.197} \\
& TAPT\_Bias      & 0.483 & 0.215 \\
& TAPT\_GN        & 0.492 & 0.219 \\
& TAPT\_GN\_Bias  & \textbf{0.493} & 0.220 \\
& TAPT\_from\_DAPT& 0.482 & 0.215 \\
& TAPT\_only      & 0.471 & 0.199 \\
\midrule
\multirow{7}{*}{\textsc{XLM-Base}}
& Baseline        & 0.516 & 0.309 \\
& DAPT\_only      & 0.522 & 0.317 \\
& TAPT\_Bias      & 0.551 & 0.349 \\
& TAPT\_GN        & 0.553 & 0.349 \\
& TAPT\_GN\_Bias  & 0.553 & 0.349 \\
& TAPT\_from\_DAPT& \textbf{0.557} & 0.356 \\
& TAPT\_only      & 0.519 & \textbf{0.313} \\
\midrule
\multirow{7}{*}{\textsc{XLM-Large}}
& Baseline        & 0.530 & 0.342 \\
& DAPT\_only      & 0.532 & 0.345 \\
& TAPT\_Bias      & 0.553 & 0.366 \\
& TAPT\_GN        & 0.554 & 0.365 \\
& TAPT\_GN\_Bias  & \textbf{0.555} & 0.367 \\
& TAPT\_from\_DAPT& 0.554 & 0.366 \\
& TAPT\_only      & 0.532 & \textbf{0.344} \\
\bottomrule
\end{tabular}
\caption{Representation analysis of allograph clusters. 
\emph{IntraCos Avg} ($\uparrow$) measures within-cluster cohesion by averaging cosine similarity between tokens and their cluster centroids. 
\emph{Nearest-InterCos Avg} ($\downarrow$) measures between-cluster separation by averaging the cosine similarity of each cluster to its nearest neighbor. 
Together, indicate how well the embedding space encodes palaeographic grapheme-allograph structure.
Best results per column are bolded.}
\label{tab:representation-all}
\end{table*}
\end{CJK} 
\end{document}